\documentclass[final]{cvpr}

\usepackage{times}
\usepackage{epsfig}
\usepackage{graphicx}
\usepackage{amsmath}
\usepackage{amssymb}

\usepackage{graphicx}
\usepackage{wasysym}
\usepackage{booktabs}
\usepackage{balance}
\usepackage{multirow}
\usepackage{overpic}
\usepackage{bbm}
\usepackage{dsfont}
\usepackage{caption}
\usepackage{color}
\usepackage{pifont}
\usepackage{etoolbox}
\usepackage{tabu}
\usepackage{xcolor}
\usepackage{bm}
\usepackage{stackengine}
\usepackage{flushend}

\newcommand{\refsec}[1]{Sec.~\ref{sec:#1}}

\newcommand{\refeq}[1]{Eq.~\ref{eq:#1}}
\newcommand{\reffig}[1]{Fig.~\ref{fig:#1}}
\newcommand{\reftab}[1]{Tab.~\ref{tab:#1}}

  \definecolor{m_Table}{RGB}{219, 206, 59}
  \definecolor{m_Chair}{RGB}{65, 80, 179}   			
  \definecolor{m_Sofa}{RGB}{221, 61, 70}  			
  \definecolor{m_Cup}{RGB}{238, 238, 238}	
  \definecolor{m_Bowl}{RGB}{81, 210, 68}  			
  \definecolor{m_Bottle}{RGB}{92, 38, 191}

\usepackage[font=small,skip=5pt]{caption}

\newcommand{\circlenum}[1]{{\textcircled{\footnotesize{$#1$}}}}

\newcommand{\papertitle}{From Points to  Multi-Object 3D Reconstruction}
 
\definecolor{darkgreen}{RGB}{0,153,51}

\newcommand{\LossCollision}{\mathcal{L}_{coll}}
\newcommand{\LossShape}{\mathcal{L}_{\mathbf{z}}}

\newcommand\ArrowDown[1]{
\hspace{-12px}\rotatebox[origin=c]{270}{$\curvearrowright$}{\hspace{2px}#1}
}

\newcommand{\legend}[2]{{\color{#1_#2}$\CIRCLE$} \hspace{-10.25px}$\ocircle$\,#2\,\,}

\usepackage[pagebackref=true,breaklinks=true,colorlinks,bookmarks=false]{hyperref}

\begin{document}

\title{\papertitle{}}

\author{Francis Engelmann$^{1, \dagger} $ \hspace{25px}
Konstantinos Rematas$^{2}$ \hspace{25px}
Bastian Leibe$^{1}$ \hspace{25px}
Vittorio Ferrari$^{2}$\\
\vspace{-8px}
\\
$^1$RWTH Aachen University \hspace{30px} $^2$Google Research\\
}


\twocolumn[{
\renewcommand\twocolumn[1][]{#1}
\maketitle
\thispagestyle{empty}
\vspace{-0.2cm}
\center
\includegraphics[width=1.0\linewidth, trim={0 0 0 0cm},clip]{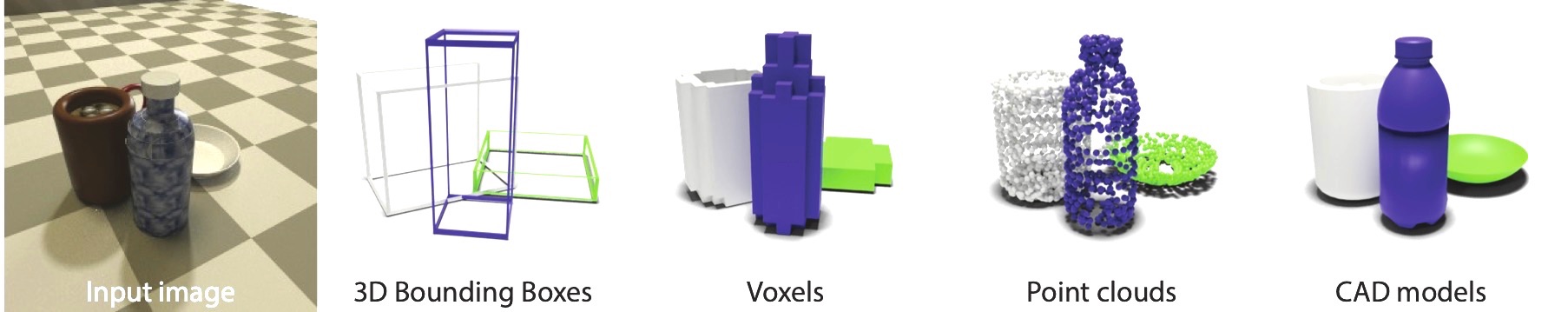}
\vspace{0px}
\captionof{figure}{
We propose a single-stage model for realistic multi-object 3D reconstruction from a single RGB image.
The model detects object center-points and performs reconstruction by jointly estimating 9-DoF bounding boxes and representation-agnostic 3D shape exemplars.}
\label{fig:teaser}
\vspace{0.5cm}
}]
\let\thefootnote\relax\footnotetext{$\dagger$ \text{Work performed during internship at Google Research, Zurich.}}

\begin{abstract}
We propose a method to detect and reconstruct multiple 3D objects from a single RGB image.
The key idea is to optimize for detection, alignment and shape jointly over all objects in the RGB image, while focusing on realistic and physically plausible reconstructions.
To this end, we propose a key-point detector that localizes objects as center points 
and directly predicts all object properties, including 9-DoF bounding boxes and 3D shapes -- all in a single forward pass.
The proposed method formulates 3D shape reconstruction as a shape selection problem, \ie it selects among exemplar shapes from a given database.
This makes it agnostic to shape representations, which enables a lightweight reconstruction of realistic and visually-pleasing shapes based on CAD-models,
while the training objective is formulated around point clouds and voxel representations.
A collision-loss promotes non-intersecting objects, further increasing the reconstruction realism.
Given the RGB image, the presented approach performs lightweight reconstruction in a single-stage, it is real-time capable, fully differentiable and end-to-end trainable.
Our experiments compare multiple approaches for 9-DoF bounding box estimation, evaluate the novel shape-selection mechanism and compare to recent methods in terms of 3D bounding box estimation and 3D shape reconstruction quality.
\end{abstract}


\begin{figure*}
    \centering
    \includegraphics[width=\textwidth]{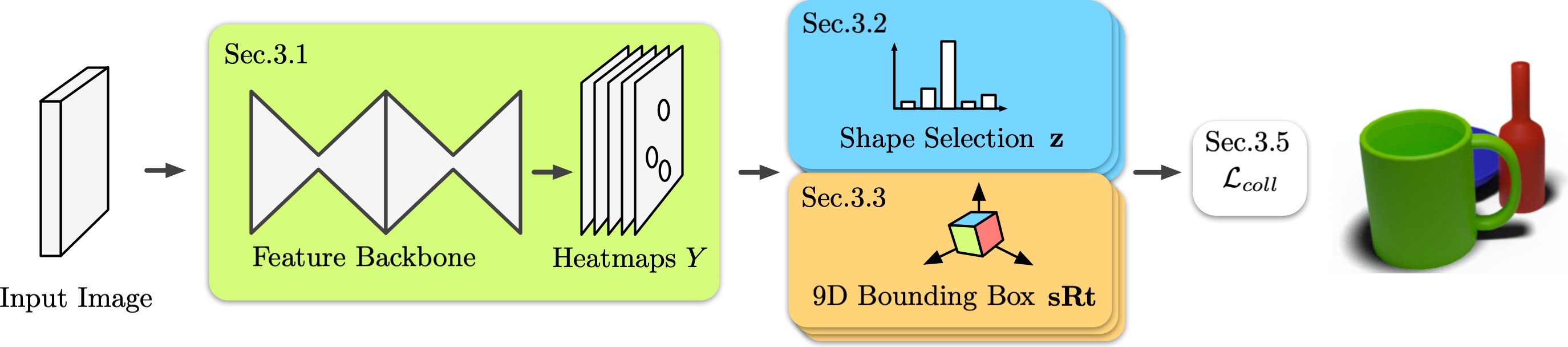}
    \caption{\small\textbf{Overview of the proposed approach.}
    Given a single RGB image, our model detects object centers as key-points in a heatmap~$Y$.
    The network directly predicts shape exemplars $\mathbf{z}$ and 9-DoF bounding boxes jointly for all objects in the scene.
    The collision loss $\mathcal{L}_{coll}$ favors non-intersecting reconstructions.
    Our method predicts lightweight, realistic and physically plausible reconstructions in a single pass.}
    \label{fig:model}
\end{figure*}

\section{Introduction}
Extracting 3D information from a single image has multiple applications in computer vision, robotics and scene understanding, specifically on mobile AR/VR devices.
Thus, this field has gained great momentum in the computer vision community~\cite{Gkioxari2019MeshR,kuo2020mask2cad,Nie20CVPR,Popov20ECCV,Jiajun2016}.
3D information can come in many forms: 3D bounding boxes, point clouds, meshes, voxels or distance fields. The choice of the representation often depends on the task.
In this paper, we aim to extract all the above information in an efficient and scalable way, all from just a single view and in a single pass.

Recent methods \cite{Gkioxari2019MeshR, kuo2020mask2cad} perform multi-object reconstruction by independently processing detections from state-of-the-art object detectors \cite{HeMaskRCNN, Kuo19ARXIV} or jointly predict multiple objects in a dense voxel grid \cite{Popov20ECCV}, which can be computationally expensive due to scalability issues.
Instead, inspired by CenterNet~\cite{Zhou19arxiv}, a framework for accurate and efficient 2D object detection, we propose to use a key-point detector to localize objects as sparse center-points and directly predict 9-DoF bounding boxes and shapes jointly for all objects in the scene. The CenterNet architecture is modular and can easily be extended to solve varying tasks such as 2D detection, 3D detection, human body pose estimation and tracking \cite{yin2020center, zhou2020tracking}.
In this paper, we argue for a complete and coherent 3D reconstruction of multiple objects using CenterNet where each pixel votes for a class label, a 3D bounding box, and a 3D shape exemplar to place objects into the world coordinate frame.

Another key question is the best shape representation.
While numerous representations have been proposed, \eg Signed Distance Functions (SDF) \cite{Park2019CVPR}, meshes \cite{Gkioxari2019MeshR, groueix2018}, voxel grids \cite{Popov20ECCV}, point clouds \cite{Fan16, Jiang2018GALGA}, and even hybrid approaches \cite{Runz2020CVPR}, all have their task-dependent advantages and disadvantages.
In this work, we propose a representation-independent shape selection mechanism.
That is, shape exemplars are selected from a given shape database that can implement different (or multiple) representations.
The most convenient representation is chosen depending on the task at hand, be it for defining objective functions or for visualization purposes (see  \reffig{teaser}).

Additionally, we take extra provisions for a realistic and physically plausible reconstruction.
In particular, objects should be properly placed in the world frame and should not intersect with each other. Inspired by recent methods on human body pose estimation in 3D scenes \cite{Hassan19ICCV, Jiang20CVPR, Zhang203DV}, we add a collision loss that supports plausible reconstructions such that reconstructed objects do not intersect.
To summarize, given a RGB image, our single-stage method performs lightweight reconstruction, it is real-time capable, fully differentiable and end-to-end trainable.
In our experiments, we compare different 9-DoF bounding box formulations,
we evaluate our shape selection mechanism using soft labels and compare with the current state-of-the-art CoReNet~\cite{Popov20ECCV}.

\vspace{-12px}
\paragraph{Contributions.}
Our key contributions are:
\begin{itemize}
    \vspace{-6px}
    \item We propose a method for multi-object 3D reconstruction that extends the CenterNet~\cite{Zhou19arxiv} framework to perform fully holistic 3D scene reconstruction in a single-stage network and from a single RGB image.
    \vspace{-6px}
    \item We present a shape-selection mechanism to perform 3D object reconstruction,
    where we reformulate the 1-of-K classification task using soft target labels based on geometric similarities between exemplar 3D shapes: this significantly improves over hard-labels as used in previous baselines~\cite{Tatarchenko19CVPR}.
    \vspace{-6px}
    \item We obtain physically plausible reconstructions by leveraging a collision loss that encourages non-intersecting reconstructions. Further, CAD based representations guarantee valid and realistic shapes.
    \vspace{-6px}
     \item Our approach is agnostic to different shape representations.
     Since we formulate the shape reconstruction problem as selecting a shape exemplar
     (\ie, index in a precomputed database of shapes),
     we can choose from any representation given the estimated shape exemplar.
\end{itemize}

\section{Related Work}
\vspace{-5px}  
\paragraph{3D from a single image.}
Single image 3D reconstruction has seen tremendous progress over the last years, with various shape representations being examined. Works like \cite{choy20163d, Girdhar16b, henzler2019platonicgan, richter2018, marrnet, Jiajun2016} operate on voxel grids, a representation that fits very well with convolutional neural networks. Other methods output point clouds \cite{Fan16, Jiang2018GALGA}, taking advantage of their compactness.
One line of work~\cite{DIBR19, groueix2018, kato2018neural, liu2019softras, wang2018pixel2mesh} outputs meshes, a powerful representation that provides neighborhood structure to the 3D shape.
Recently, implicit representations \cite{chen_cvpr19,Mescheder2019CVPR,Niemeyer2020CVPR,Park_2019_CVPR} have gained popularity for their ability to represent fine details at arbitrary resolutions. An alternative to the 3D shape regression is the work of \cite{Tatarchenko19CVPR} that poses the 3D reconstruction as a classification/retrieval problem. 
However, all of these methods focus on the single object case: the image contains a single object to be reconstructed, often on a white background. By having every pixel predict a 3D bounding box, a shape index similar to \cite{Tatarchenko19CVPR}, and the 9-DoF, we are able to handle arbitrary number of objects in the scene and in a single forward pass.

\vspace{-15px}
\paragraph{Multi-object 3D reconstruction.}
Recently, multi-object 3D reconstruction made significant progress: Im2CAD \cite{izadinia2017im2cad} performs object detection and room layout estimation in an input image, and then retrieves 3D shapes from a database and aligns them to match the detections. However, it involves a secondary non-differentiable optimization step, that renders and matches the estimations with the input image.
3D-RCNN~\cite{3DRCNN_CVPR18} estimates the 3D shape of each object instance in an image through a render-and-compare learning approach, where the shape is represented as a linear basis from a dataset of 3D models. This shape representation though is accurate for classes with low intra-class variability such as cars and humans.
Given an image and a set of object proposals,~\cite{factored3dTulsiani17} decompose the underlying 3D scene into a room layout, a set of voxels grids for every object, together with their rotation/translation/scaling parameters.
Similarly, ~\cite{Nie20CVPR,zhang2021holistic} propose to estimate the room layout, 3D object bounding boxes and shape for every object. However, the 3D estimates depend on the initial 2D bounding boxes.
\cite{Avetisyan20ECCV,griffiths2020finding} make use of center predictions but require 3D reconstructions as input.
In the work of~\cite{huang2018holistic}, the 3D scene is represented as a graph that is being optimized so the configuration of objects and room layout matches the semantic and geometric properties of the input image. Mesh R-CNN~\cite{Gkioxari2019MeshR} can be seen as an extension of Mask-RCNN~\cite{HeMaskRCNN} to estimate 3D meshes for every object instance in an image, but without resolving their scale/depth ambiguity.

Mask2CAD~\cite{kuo2020mask2cad} shares with our work the elements of center prediction and CAD model retrieval. The main differences are:
(1) we base our architecture on CenterNet (\vs ShapeMask) leading to a simpler model that can be trained end-to-end more easily;
(2) we predict a complete 9-DoF pose, whereas \cite{kuo2020mask2cad} requires given object depth at test time, and returns the object scaled as in the database (instead, we can stretch it along each of the 3 dimensions);
(3) we include a  collision loss dedicated to improving estimation of nearby objects.
(4) we directly predict pose as a valid rotation matrix (\vs two-stage approach).

The above works are based on complex, two-step architectures, first detecting objects and then estimating their shape.
In contrast, our method is single-step, scales well with the number of objects in an image, and does not involve post-processing mechanisms. 

\vspace{-10px}
\paragraph{CoReNet}~\cite{Popov20ECCV} performs dense shape prediction in a fixed $128^3$ voxel grid, which does not scale with the size of the reconstructed world.
Moreover, it bakes all scene information into one model during training (number of objects, class combinations).
Instead, our approach is more modular, it can detect and reconstruct a variable number of objects, as well as new combinations of classes not seen during training.
Our approach predicts both a 9-DoF oriented bounding box and shape.
Additionally, our shape representation is independent of the actual representation. We can predict signed distance functions, point clouds, occupancy grids and meshes, which naturally leads to realistic scene reconstructions,
whereas CoReNet tends to predict holes/errors, especially in multi-object scenes.


\begin{figure}[t]
    \centering
    \frame{\includegraphics[width=0.49\columnwidth]{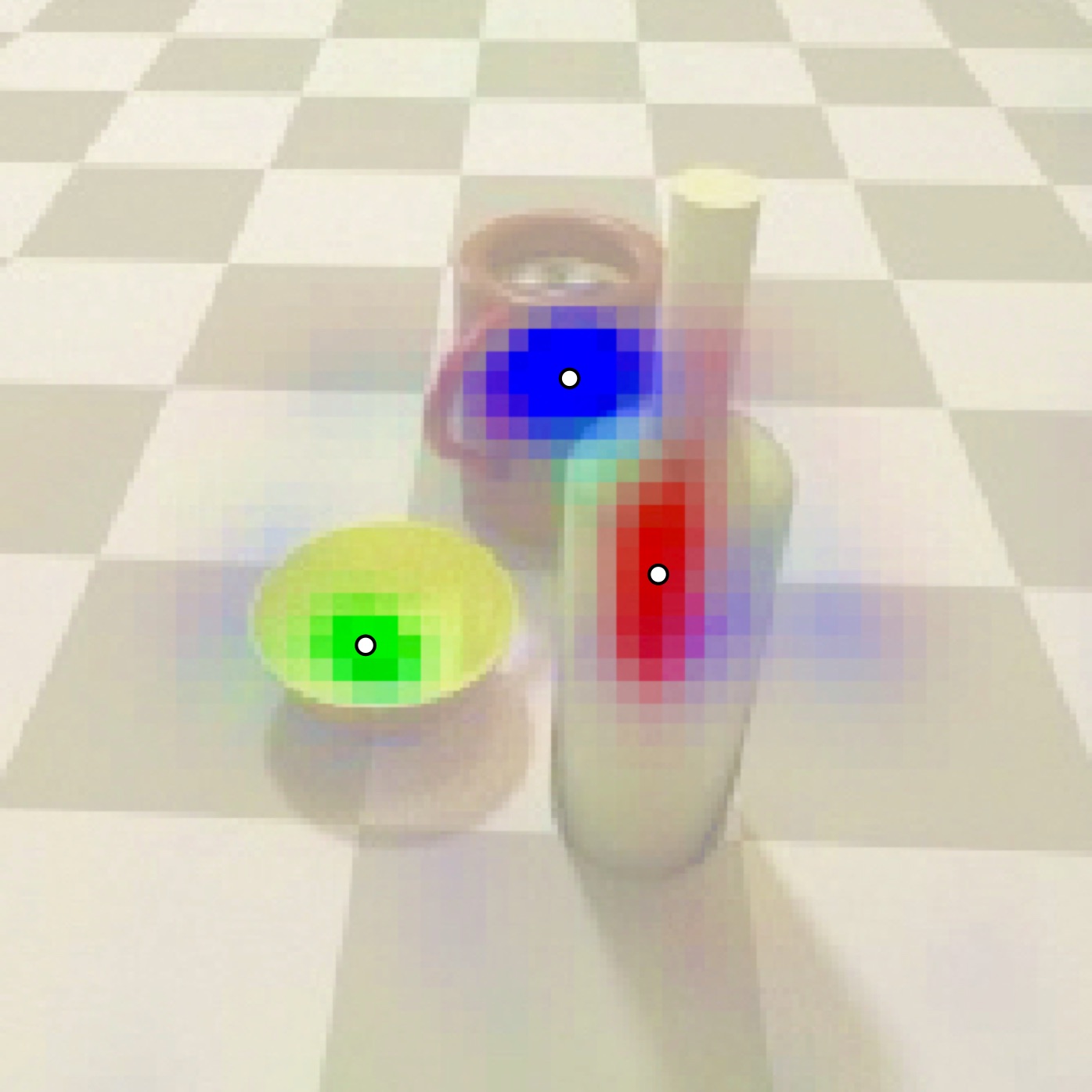}}
    \hfill
    \frame{\includegraphics[width=0.49\columnwidth]{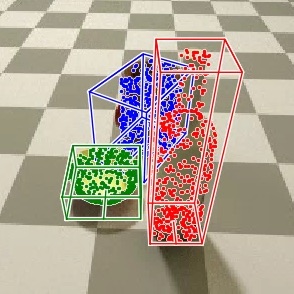}}
    \caption{\small\textbf{Object detection as key-point detection.}
    \emph{Left:}
    Predicted heatmaps $\hat{Y}$ visualizing the per-pixel probability for being an object center.
    The heatmap $\hat{Y}_c$ of each class $c$ is shown in a different color. The peaks of the distributions are shown as white circles $\circ$, they correspond to the detected object centers\,$\hat{\mathbf{p}}$ from which the object properties are predicted.
    \emph{Right:}
    Predicted object properties.
    We show the estimated 9-DoF bounding boxes and the 3D shapes using the point cloud representation.
    }
    \label{fig:detection}
\end{figure}

\section{Method Overview}
This section introduces each module and the corresponding losses of our full model shown in \reffig{model}.
We formulate object detection as a key-point detection problem similar to \emph{CenterNet}~\cite{Zhou19arxiv},
where each object is represented by its center point in the 2D image (\refsec{key_point_detection}).
From the detected center points, we directly estimate
realistic shapes (\refsec{shape_reconstruction})
and oriented 3D object bounding boxes (\refsec{3d_bounding_box}).
To further promote physically plausible reconstructions, we propose a collision loss to avoid intersecting objects (\refsec{collision_loss}).

\subsection{Object Detection as Key-Point Detection}
\label{sec:key_point_detection}

The first part of our method is a key-point detector that follows the setup of \emph{CenterNet}~\cite{Zhou19arxiv}.
Given a single RGB image $I \in \mathbb{R}^{W \times H \times 3}$,
the detector localizes key-points (here: object centers) by predicting class-specific heatmaps $\hat{Y} \in [0,1]^{\frac{W}{R} \times \frac{H}{R} \times C}$
(\reffig{detection}, \emph{left})
where $C$ is the number of object classes and $R$\,=\,$4$ is a down-sampling factor.
The detected center points $\{\hat{\mathbf{p}}_i \in \mathbb{R}^2 \}$ (shown as $\circ$ in \reffig{detection}) correspond to the local maxima in the predicted heatmaps $\hat{Y}$.
They are obtained using non-maximum-suppression, which is implemented as a $3 \times 3$ max pooling.
We associate a confidence score $s_i = \hat{Y}_{\hat{\mathbf{p}}_i}$ to each detected key-point $\hat{\mathbf{p}}$.
The feature backbone -- which takes the input image $I$ and generates the output heatmaps $\hat{Y}$ --
is implemented as a stacked hourglass model~\cite{Newell2016ECCV}.

During training, we follow \cite{Law2018ECCV, Zhou19arxiv} and generate the target heatmaps $Y$ by splatting the ground truth center points $\mathbf{p}_i$ using Gaussian kernels $\mathcal{N}(\mathbf{p}_i, \sigma_i)$ with $\sigma_i$ depending on the projected size of the object $i$.
Training the key-point detector relies on the focal loss~\cite{Lin17ICCV} and is computed over all pixels $(x, y)$ and classes $c \in \{1, \ldots, C\}$ in the heatmaps:
\begin{equation}
    \mathcal{L}_{key}\text{\,=\,}\frac{-1}{N} \sum_{xyc} \begin{cases}
      (1\text{\,-\,}\hat{Y}_{xyc})^{\alpha} \cdot \text{log}(\hat{Y}_{xyc}) &\hspace{-25px}\text{if $Y_{xyc}$\,=\,1}\\
      (1\text{\,-\,}Y_{xyc})^{\beta}  \cdot  (\hat{Y}_{xyc})^{\alpha} \cdot \text{log}(1\text{\,-\,}\hat{Y}_{xyc}) &\hspace{-0px}\text{else}\\
    \end{cases}
\end{equation}
where $N$ is the number of ground truth objects, $\alpha$\,=\,2 and $\beta$\,=\,4 are the hyper-parameters of the focal loss.
After detecting the object instances as center points, the network jointly
selects 3D shapes (\refsec{shape_reconstruction}) and 
estimates 3D bounding boxes (\refsec{3d_bounding_box})
for each object in the scene.


\begin{figure}
    \centering
    \includegraphics[width=0.95\columnwidth]{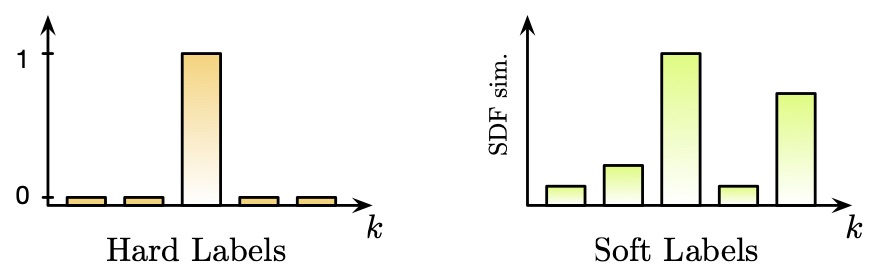}
    \caption{\small\textbf{Shape selection.}
    We compare one-hot encoding (hard labels, \emph{left}) for supervising the shape selection problem with soft labels (\emph{right}) which allow for multiple shape predictions at the same time and are based on geometric similarity, specifically the Euclidian distance between SDF shape representations.}
    \label{fig:shape_selection}
\end{figure}

\subsection{Shape Selection}
\label{sec:shape_reconstruction}
Instead of directly reconstructing shape representations such as meshes, voxel grids or point clouds~\cite{Fan16,groueix2018,Popov20ECCV},
our method operates indirectly, by selecting shape exemplars.
More precisely, the network is trained to select for each detection one shape exemplar $z$
among a set of $K$ shape exemplars from a given shape database.
This choice is motivated by our goal to reconstruct realistic scenes, since it guarantees valid shapes from the object database unlike recent reconstruction methods which can produce incomplete, noisy or over-smoothed reconstructions.
Similarly, the recent work of Tatarchenko \etal~\cite{Tatarchenko19CVPR} concludes that current methods for single-view 3D reconstruction primarily work because of recognizing the type of shape depicted in the image, rather than truly recovering the geometric details unique to that particular instance.

To reiterate, in this work, the shape estimation problem is formulated as a shape selection problem which chooses one shape exemplar $\hat{z}$ from a given shape database $\mathcal{Z}$ of $K$ shape exemplars.
After predicting an exemplar $\hat{z}$, an explicit shape representation $X$ (voxel-grid, point cloud, CAD model \etc) can be chosen freely from the precomputed databases $\mathcal{Z}_X$ (described next) depending on the task or loss function at hand.
As such, the presented model is agnostic towards any particular shape representation.

\paragraph{Building the shape database $\mathcal{Z}$.}
The presented shape database is a set of representative shape exemplars selected from a given set of CAD models.
Once our shape database is built, the full set of the original CAD models is no longer required.
We now describe how those exemplary shapes are selected.
First, the CAD models are transformed into a canonical orientation, position and scale.
Specifically, all models are facing down the negative Z-axis, the centroids are translated to the origin, and we apply anisotropic scaling such that the models fit into the unit cube. 
Then, for each object $i$, we compute the signed distance function (SDF) representation  $\phi^i$ of the corresponding CAD model.
After discretization, downsampling to $32^3$ grids and flattening to vectors, we cluster the objects using k-Means++~\cite{David07ACM} with $k$\,=\,$50$, for each object class separately.
The total number $K$ of shape exemplars in the database $\mathcal{Z}$ is $K = k \cdot C$ where $C$ is the number of object types (chairs, bottle, \etc). 
The objects appearing in the training images are already annotated by their corresponding CAD model.
Hence, we can re-label each object with their nearest shape exemplar $z^k$.
Additionally, the shape database can be extended to store explicit shape representations such as SDFs
$\mathcal{Z}_{\phi}$\,=\,$\{\phi^k\}_{k=1}^K $, point clouds $\mathcal{Z}_{\mathcal{P}}$\,=\,$\{\mathcal{P}^k\}_{k=1}^K$ or CAD models $\mathcal{Z}_{\text{CAD}}$\,=\,$\{\text{CAD}^k\}_{k=1}^K$.
In each case, the stored representation corresponds to the model that is closest to the cluster center under the clustering metric ($L_2$ distance over $\phi$).

\paragraph{Training the shape selection network module.}
One straightforward approach consists in training a 1-of-$K$ classifier.
Specifically, for each object $i$ in the input image, the network predicts a vector $\hat{{\mathbf{z}}}^i \in \mathbb{R}^{K}$
scoring it against each of the $K$ exemplar shapes in the shape database $\mathcal{Z}$.
We can then place a cross-entropy loss $CE(\cdot, \cdot)$ on this output and supervise it with the ground truth one-hot encoding of the target shape $\mathbf{z}^i \in \{0, 1\}^{K}$ (\reffig{shape_selection}, \emph{left}):
\begin{align}
\mathcal{L}'_{\mathbf{z}} &= \frac{1}{M} \sum_{i=1}^M CE\big(\mathbf{z}^i, \sigma{(\hat{\mathbf{z}}^i)}\big) \\
								&= -\frac{1}{M} \sum_{i=1}^M \sum_{k=1}^K {z}^i_k \cdot \text{log}\big(\sigma(\hat{{\mathbf{z}}}^i)_k\big)
\label{eq:shape_hard}
\end{align}
where $M$ is the number of detections in the image, $\sigma$ is the softmax function (\cf next paragraph, where we use sigmoid $S$ instead), and $z^i_k$ is the $k$-th entry in vector $\mathbf{z}^i$.
At test time, the predicted shape exemplar $\hat{z}^i$ is computed as $\hat{z}^i = \text{argmax}_k(\hat{\mathbf{z}}^i)$.
This approach corresponds to the clustering baseline presented by Tatarchenko \etal in \cite{Tatarchenko19CVPR}.

The issue with this approach is that two objects $\{i, j\}$ that are geometrically similar (\ie $\phi^i \approx \phi^j$) can have disagreeing supervision signals $\{\mathbf{z}^{i}, \mathbf{z}^{j}\}$.
This can have a negative impact on the network training, as the network is asked to simultaneously predict a high value for one of the $K$ database shapes, while also predicting a low value for another, very similar shape.
Instead, we propose as alternative formulation a soft relaxation of the binary target labels $\mathbf{z} \in \{0, 1\}^K$ which takes the geometric similarity of shapes into account.
Specifically, we allow to predict multiple shape exemplars simultaneously, they are no longer mutually exclusive as before.

Formally, we redefine the target labels $\mathbf{z}$ using a shape similarity function $d(\cdot, \cdot)$ (\reffig{shape_selection}, \emph{right}) such that:
 \begin{equation}
  \mathcal{L}_{\mathbf{z}} = - \frac{1}{M} \sum_{i=1}^M \sum_{k=1}^K d(i, k) \cdot \text{log}\big(S(\hat{z}_k)\big)
  \label{eq:shape_soft}
 \end{equation}
where $S$ is the sigmoid function and
\begin{equation}
        d(i, k) = [1 - \lVert \phi^i  - \phi^k \rVert_2 ]_{+}
\end{equation}
where $[\,\cdot\,]_+ = \text{max}(\,\cdot\,,\,0)$ and $\lVert \cdot \rVert_2$ is the Euclidean distance between the shape exemplars' SDFs $\phi^k$ in the shape database $\mathcal{Z_{\phi}}$, and $\phi^i$ is the ground truth SDF of object $i$.

In the following, we will refer to these labels as \emph{soft}-labels, and when using the one-hot encoding as \emph{hard}-labels.
In \refsec{experiments}, we show that this alternative soft formulation is key to improve shape selection.
At test time, we simply select the shape exemplar with the highest output value by the network.
Next, we describe our approach to estimate the 3D bounding boxes,
which are subsequently used to transform the estimated object shapes from their canonical database pose into the scene coordinate frame.

\subsection{3D Bounding Box Estimation (9-DoF Poses)}
\label{sec:3d_bounding_box}
Along with the realistic shape representation we aim at finding a 9-DoF bounding box for each object in the input image $I$.
We describe now the estimation of the 9-DoF bounding box parameters, capturing the object pose in the scene.
They include a 3D rotation $\hat{\mathbf{R}} \in SO(3)$, a 3D translation $\hat{\mathbf{t}} \in \mathbb{R}^3$ and a 3D scale $\hat{\mathbf{s}} \in \mathbb{R}^3$.
These parameters are used to transform the estimated object shape from its canonical database pose to the scene coordinate frame.

In \emph{CenterNet}, Zhou \etal~\cite{Zhou19arxiv} formulate the rotation estimation as a combination of classification over quantized bins followed by regression to a continuous offset.
That formulation requires the definition of multiple loss functions along with carefully tuned loss weights.
Instead, we directly parameterize the object rotation as a 3D rotation matrix $\hat{\mathbf{R}} \in SO(3)$.
Specifically, our network predicts a 9-dimensional output interpreted as a $3\times3$ rotation matrix $\mathbf{M}$ with (differentiable) SVD decomposition~\cite{Gene1996} $\mathbf{M}$\,=\,$\mathbf{U\Sigma V}^{\top}$.
The corresponding symmetric orthogonal rotation matrix $\hat{\mathbf{R}}$ is then obtained by projecting $\mathbf{M}$ into $SO(3)$~\cite{Levinson20arxiv}:
\begin{equation}
	\hat{\mathbf{R}} = \mathbf{U} \mathbf{\Sigma}' \mathbf{V}^{\top}, \; \text{where } \mathbf{\Sigma}' = \text{diag}\big([1, 1, \text{det}(\mathbf{U}\mathbf{V}^{\top})]\big)
\end{equation}
While more straightforward, this formulation can directly be optimized using, \eg, the Frobenius norm~\cite{Gene1996}: $\lVert \mathbf{R} - \hat{\mathbf{R}} \rVert_F$.
The translation $\hat{\mathbf{t}} \in \mathbb{R}^3$ is defined as the vector from the scene origin to the 3D bounding box centroid,
and can be optimized, \eg, with the \emph{Huber} loss (smooth-$L_1$): $\lVert \mathbf{t} - \hat{\mathbf{t}} \rVert_H$.
Instead, we propose to jointly optimize both the rotation $\hat{\mathbf{R}}$ and the translation $\hat{\mathbf{t}}$ using the concatenated transformation $\mathbf{T} = [\mathbf{R}\,| \,\mathbf{t}]$.
Specifically, we minimize the squared Euclidean distance between the point cloud $\mathcal{P}^i$ of the object under the estimated $\hat{\mathbf{T}}$ and ground truth transformation $\mathbf{T}$.
Formally, we have:
\begin{equation}
	\mathcal{L}_{\mathbf{Rt}} = \sum_{i=1}^M \sum_{\mathbf{x} \in \mathcal{P}^i} \lVert \mathbf{T}^i\,\mathbf{x} - \hat{\mathbf{T}}^i\,\mathbf{x} \rVert_2^2
	\label{eq:pose}
\end{equation}
where $M$ is the number of objects in the image, $\mathbf{x} \in \mathbb{R}^3$ is a point in the point cloud $\mathcal{P}^i$ sampled from the surface of the ground truth object $i$ in the input image.

Finally, the scale loss $\mathcal{L}_{\mathbf{s}}$ is implemented as the $\text{L}_1$ distance between predicted and ground truth 3D scale averaged over all objects in the input image.
Similar to \cite{Zhou19arxiv}, the neural network branch that predicts the bounding box parameters is class-agnostic (\ie the same for all classes $c$) and only receives supervision at the ground truth center locations.
In summary, the loss for the 9-DoF bounding box estimation consists of two terms:
$\lambda_{Rt}\mathcal{L}_{Rt} +\lambda_{s}\mathcal{L}_{s}$.


\begin{figure}
    \centering
	\stackinset{l}{0.2cm}{t}{0.6cm}{CAD$^j$}{\stackinset{l}{3.0cm}{t}{3.0cm}{CAD$^i$}{\frame{\includegraphics[width=0.49\columnwidth]{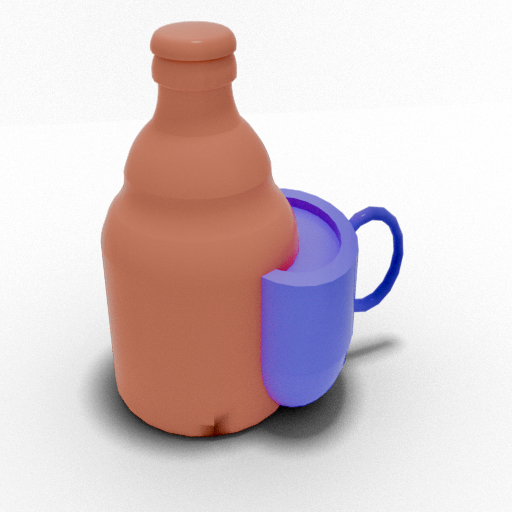}}}}%
    \hfill
	\stackinset{l}{0.6cm}{t}{0.6cm}{$\widetilde\phi^j$}{\stackinset{l}{3.0cm}{t}{2.8cm}{$\mathcal{P}^i$}{%
	\frame{\includegraphics[width=0.49\columnwidth]{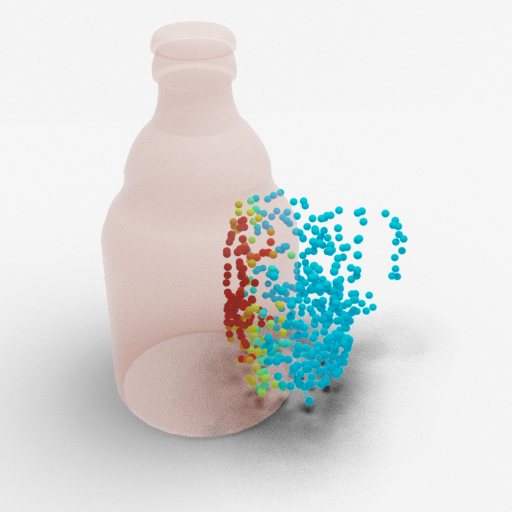}}}}%
    \caption{\small\textbf{Visualization of the collision loss.}
    The collision loss penalizes colliding objects, contributing to an improved realism of the reconstructed scene. 
    \emph{Left:} Physically implausible reconstruction of two colliding objects.
    \emph{Right:} The colors represent the SDF values sampled at the point positions $\mathcal{P}^i$ of the cup in the SDF $\widetilde \phi^j$ of the bottle.
    Outside the object the sampled values are zero (blue) and increase with the distance to the surface (from blue to red).}
    \label{fig:collision_illustration}
\end{figure}

\subsection{Collision Loss}
\label{sec:collision_loss}

Towards our goal of realistic multi-object reconstruction,
it is not only important that the individual objects exhibit realistic shapes,
but also that their poses form a physically plausible spatial configuration in the scene.
One specific concern is that reconstructed objects should not intersect or collide 
with each other.
However, the model we just presented in practice often predicts colliding shapes, especially for nearby objects.

As a remedy, we propose to add a collision loss that inflicts a penalty whenever two or more reconstructed objects collide.
In particular, we rely on the convenient property of our model that it can choose from multiple shape representations
and use the SDF representation $\phi^j$ of an object $j$ and the point cloud $\mathcal{P}^i$ of another object $i$ to compute the point-to-surface distance.
Specifically, the SDF reveals $\phi^j$ the distance of a point to the nearest surface of object $j$. It is negative inside the object and positive outside. Therefore, we define $\widetilde \phi = \text{min}(-\phi, 0)$ such that the values are positive inside the object and zero outside.
Formally, the collision loss for one object $i$ with all other objects $j$ is:
\begin{equation}
    \LossCollision^{i} = \sum\limits_{\substack{j = 1 \\ i\neq j}}^M \sum_{\mathbf{x} \in \mathcal{P}^i}
    \widetilde{\phi}^j(\mathbf{T}^{ij} \mathbf{x})
\end{equation}
where $M$ is the total number of detections in the scene,
$\mathbf{T}^{ij}$ is the transformation matrix 
placing the point cloud $\mathcal{P}^i$ of object $i$ into the local coordinate system of object $j$.
As we store the SDFs values as discrete voxel grids, we perform differentiable trilinear interpolation when sampling $\widetilde\phi^j$ at the continuous point positions $ \mathbf{T}^{ij} \mathcal{P}^i$.  
\reffig{collision_illustration} provides a visual interpretation of the loss. Inside object $j$, the SDF $\widetilde\phi^j$ is positive and zero outside.
Note that the SDF $\phi$ and point clouds $\mathcal{P}$ can be pre-computed, as the shape reconstruction task is formulated as an exemplar selection problem in our model, so all possible output shapes are known beforehand.

The collision loss over all objects in a scene is:
\begin{equation}
    \LossCollision = \sum_{i = 1}^M \rho ( \LossCollision^i )
\end{equation}
where $\rho(x)$\,=\,$\frac{x^2/2}{1 + x^2}$ is the robust Geman-McClure loss \cite{Geman87Statistical} compensating for varying point densities among objects.

\subsection{Training Details}

\begin{figure}[h!]
    \centering
    \includegraphics[width=\columnwidth]{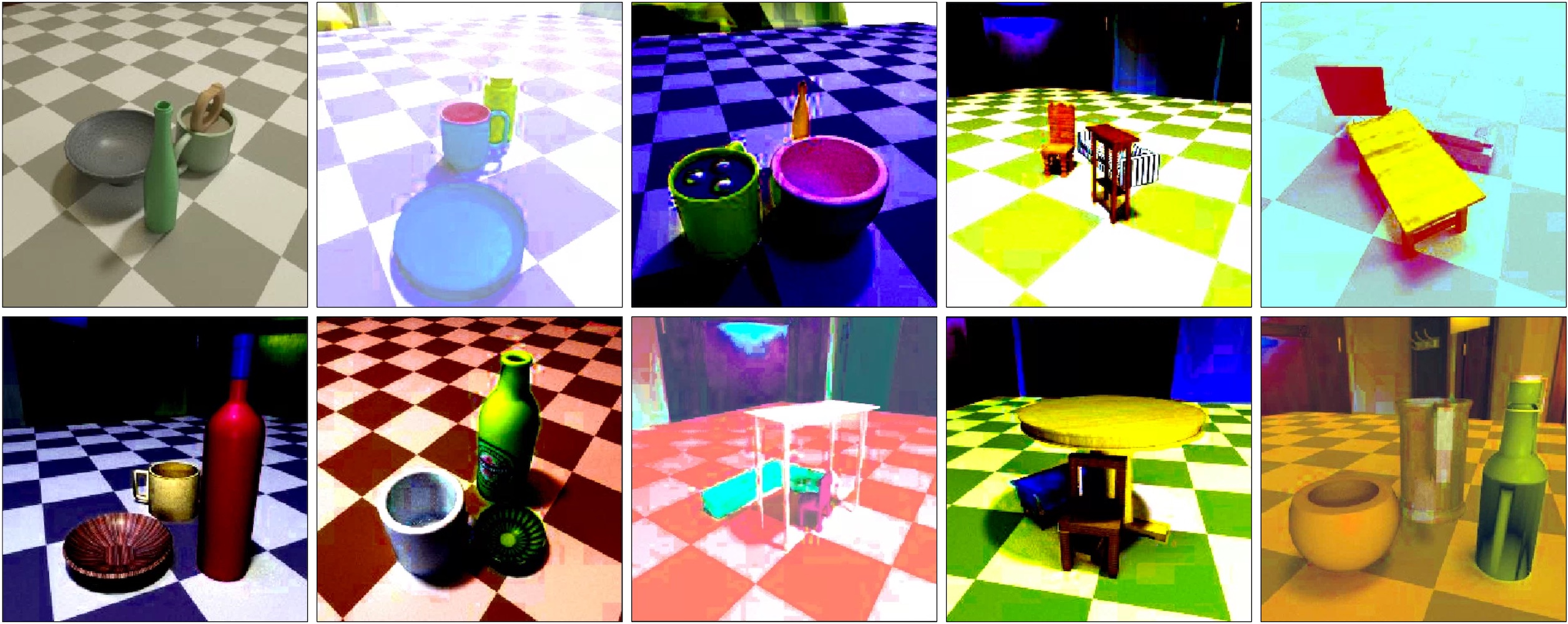}
    \caption{\small\textbf{Data augmentation examples}. Strong data augmentation is essential.
    We perform HSV-color augmentation and random horizontal flipping.
    For comparison, the top-left image shows an example that is not augmented.}
    \label{fig:data_augmentation}
\end{figure}

The full model is optimized by minimizing the multi-task loss $\mathcal{L}$ defined using the previously introduced losses:
\begin{equation} 
    \mathcal{L} = \mathcal{L}_{key} + 
    \lambda_{\mathbf{Rt}}\mathcal{L}_{\mathbf{Rt}} + 
    \lambda_{\mathbf{s}}\mathcal{L}_{\mathbf{s}} + \lambda_{\mathbf{z}}\LossShape + \lambda_{coll}\LossCollision
\end{equation}
where $\lambda$ are weighting coefficients with associated values $\{10,\,10,\,0.1,\,1.0\}$ respectively.
One important observation is that the collision loss can contradict the pose losses $\mathcal{L}_{Rt}, \mathcal{L}_{s}$, especially in the beginning of the training process when the initial object pose estimates are still quite far away from the ground truth.
Penalizing colliding objects at this stage is not helpful and even has a negative impact on convergence speed.
Therefore, we enable the collision loss only after 100 epochs; before that we set its weight $\lambda_{coll} = 0$.
We train the entire network from scratch and end-to-end using the Adam optimizer,
and a batch size of 32 for 300 epochs on four P100 GPUs.
Training the model to convergence takes about 48 hours.
After 5 epochs of warm-up, we use a constant learning rate of $10^{-3}$ and perform cosine-decay after 200 epochs.
We implemented our model in TensorFlow\,2.
We found strong data augmentation to be critical for training stability.
Specifically, we perform HSV-color augmentation and random horizontal image flipping (\reffig{data_augmentation}).%

\begin{figure*}
\vspace{10px}
    \centering
	
	\includegraphics[width=0.98\textwidth,trim={0 5mm 0 0},clip]{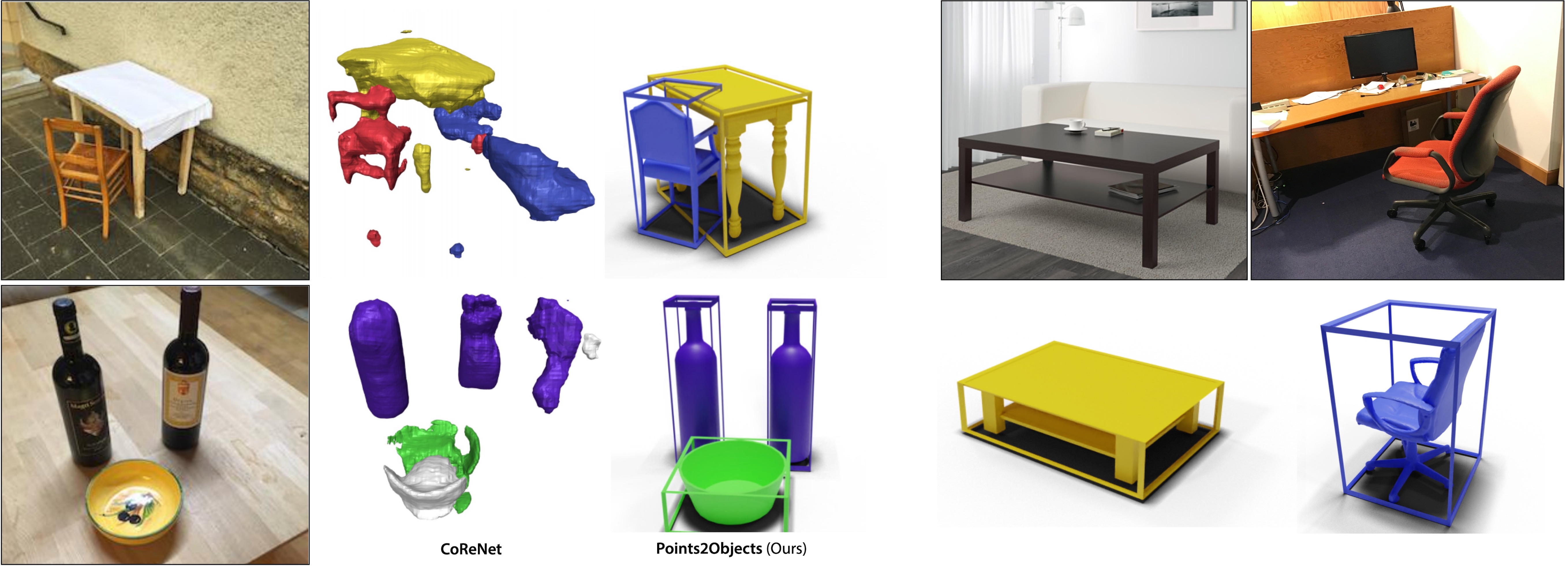}%

    \caption{\normalsize\textbf{Qualitative results on real images.}
    \emph{Left:} Comparison of our Points2Objects \vs CoReNet\,\cite{Popov20ECCV} on real images we acquired in the wild.
    \emph{Right:} Qualitative results of Points2Objects on real images from the single-object Pix3D dataset\,\cite{Sun18CVPR}.}
    \label{fig:qualitative_results_1}
\end{figure*}

\begin{figure*}
\vspace{10px}
    \centering
	
	\vspace{2mm}
	
    \frame{\includegraphics[width=0.196\textwidth]{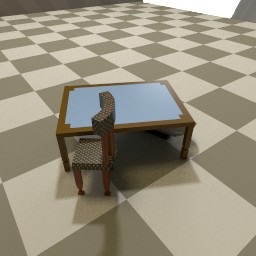}}%
    \frame{\includegraphics[width=0.196\textwidth]{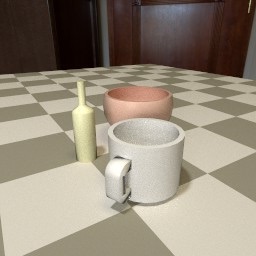}}%
    \frame{\includegraphics[width=0.196\textwidth]{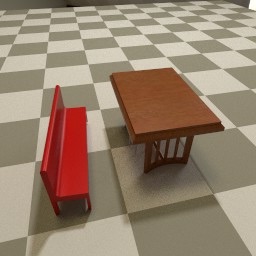}}%
    \frame{\includegraphics[width=0.196\textwidth]{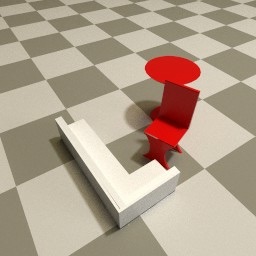}}%
    \frame{\includegraphics[width=0.196\textwidth]{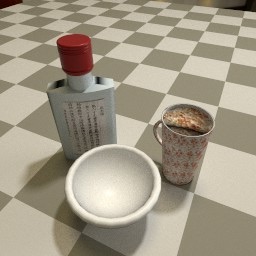}}%
    
    {\includegraphics[width=0.196\textwidth]{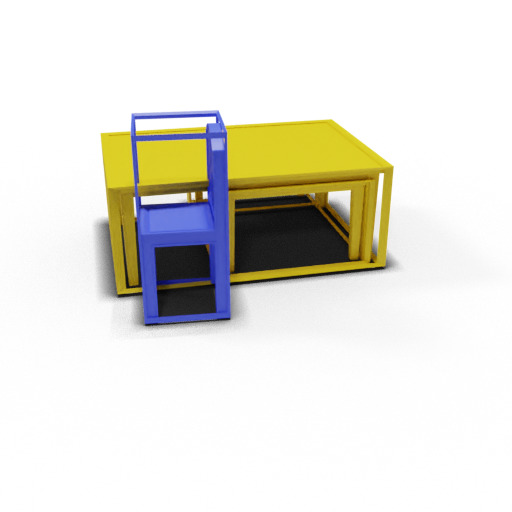}}%
    {\includegraphics[width=0.196\textwidth]{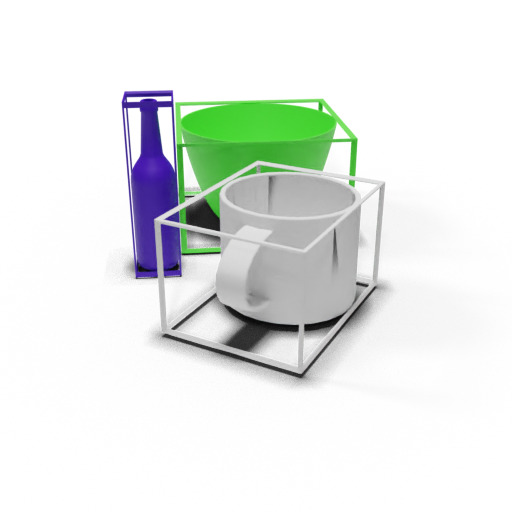}}%
    {\includegraphics[width=0.196\textwidth]{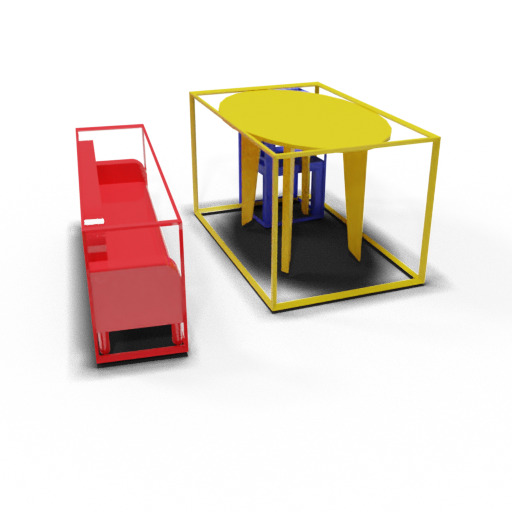}}%
    {\includegraphics[width=0.196\textwidth]{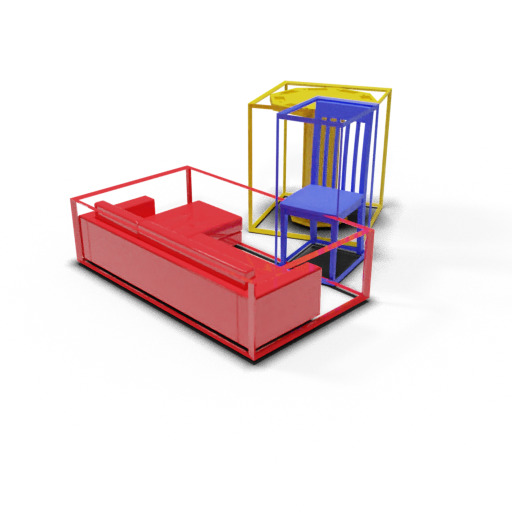}}%
    {\includegraphics[width=0.196\textwidth]{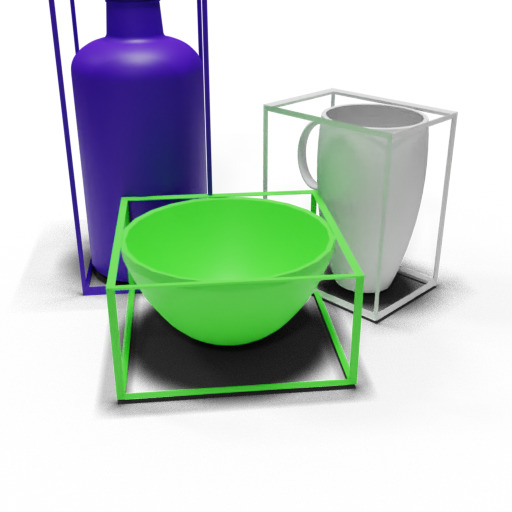}}%
    \vspace{-20px}
    \legend{m}{Table}
    \legend{m}{Chair}
    \legend{m}{Sofa}
    \legend{m}{Cup}
    \legend{m}{Bowl}
    \legend{m}{Bottle}
    \vspace{5px}
    \caption{\normalsize\textbf{Qualitative results on \cite{Popov20ECCV}.}
    \emph{Top row:} Single RGB input images.
    \emph{Bottom row:} Outputs of our method. We show the 9-DoF object bounding boxes and the selected shapes exemplars from the CAD model database $\mathcal{Z}_{\text{CAD}}$.}
    \label{fig:qualitative_results}
\end{figure*}


\begin{table*}
\begin{center}
\begin{tabular}{lcccccc c cc c cc}
\\
\toprule
& \multicolumn{6}{c}{Abs. 3D IoU per Object Class} & & \multicolumn{2}{c}{Abs. 3D IoU} & & \multicolumn{2}{c}{Rel. 3D IoU}\\
\cmidrule(r){2-7}
\cmidrule(r){9-10}
\cmidrule(r){12-13}
    & bottle     & bowl     & chair        & mug      & sofa   & table & & mean & global & & mean & global\\
\midrule
\hspace{-5px}\circlenum{1} CoReNet $m_8$ \cite{Popov20ECCV} & 61.0 & 32.2 & 30.2 & 46.8 & 54.4 & 32.4 & & 43.0 & 49.1 &&  43.0 & 49.1 \\
\hspace{-5px}\circlenum{2} CoReNet $m_9$ \cite{Popov20ECCV} & 61.8 & 36.2 & 30.1 & 48.0 & 52.9 & 34.8 & & \textbf{43.9} & \textbf{49.8} && 43.9 & 49.8  \\
\midrule
\hspace{-5px}\circlenum{3} Points2Objects \footnotesize{(Ours)}  & 63.5 & 30.2 & 18.9 & 41.5 & 44.5 & 19.8 & & 36.4 & 44.7 && \textbf{59.5} & \textbf{73.0}\\
\hspace{-5px}\circlenum{4} Points2Objects \footnotesize{(Ours, aligned)}\hspace{-10px}   & 78.2 & 39.9 & 30.6 & 47.3 & 54.9 & 38.7 & & \textbf{48.3} & \textbf{52.0} && \textbf{78.9} & \textbf{84.9}\\
\midrule
\hspace{-5px}\circlenum{5} \textit{Oracle} & \textit{86.0} & \textit{56.5} & \textit{42.1} & \textit{66.1} & \textit{66.3} & \textit{50.2} & & \textit{61.2} & \textit{61.2} &&   \textit{100} & \textit{100} \\
\bottomrule
\end{tabular}
\end{center}
\vspace{-5px}
\caption{\normalsize\textbf{Comparision with CoReNet~\cite{Popov20ECCV}}. Per-class and mean IoU over all classes and class-agnostic global IoU on $128^3$ voxel grid.
We show absolute reconstruction scores (Abs. 3D IoU) and relative scores (Rel. 3D IoU), that is, relative to the maximum possible scores.
For our model, the maximum possible score is indicated by the ground truth oracle \circlenum{5}.
}
\label{tab:corenet_triplet}
\end{table*}

\section{Experiments}
\label{sec:experiments}

We structure our quantitative evaluation in 3 parts, each addressing a core contribution of the paper:
(1) we compare multiple 9-DoF bounding box estimation mechanisms and report improved scores over the one used in CenterNet~\cite{Zhou19arxiv};
(2) the collision loss reduces the number of collisions which increases the realism and physical plausibility of the reconstructions;
(3) we show that our shape selection mechanism using soft-labels improves over hard-labels as used by~\cite{Tatarchenko19CVPR}.
Finally, we compare our method to the current state-of-the-art approach for multi-object reconstruction CoReNet~\cite{Popov20ECCV}.
\reffig{qualitative_results_1} and \reffig{qualitative_results} show qualitative results. 

\vspace{-5px}
\paragraph{Datasets}
We evaluate multi-object reconstruction using \emph{ShapeNet-pairs} and \emph{ShapeNet-triplets} datasets from~\cite{Popov20ECCV}.
They contain $256$\,$\times$\,$256$\,px photorealistic renderings of either pairs or triplets of ShapeNet\,\cite{Chang2015ARXIV} objects
placed on a ground plane with full global illumination on an environment map background, using the PBRT\,\cite{Matt16} renderer.
The scenes are rendered from a random camera viewpoint (yaw and pitch).
Objects are placed at random locations on the ground plane, with random scale, rotation, and without overlap.
This is well suited to evaluate the physical plausibility of multi-object reconstruction.
We build the shape database $\mathcal{Z}$ using ShapeNet~\cite{Chang2015ARXIV}, as the correspondences between its CAD models and the objects rendered in the images are readily available in the datasets of~\cite{Popov20ECCV}.
We set $k$\,=\,$50$, with the number of object types $C$\,=\,$6$ (ShapeNet-triplets) or $C$\,=\,$13$ (ShapeNet-pairs).
Finally, in the last part of this section we also report an evaluation on real images from the (single-object) dataset \emph{Pix3D}~\cite{Sun18CVPR}.

\vspace{-8px}
\paragraph{How to estimate 3D bounding boxes?}
We compare here the different approaches to estimate the rotation and translation of 3D bounding boxes.
Specifically, we compare the combined loss $\mathcal{L}$ from \refeq{pose} with the individual losses $\mathcal{L}_{\textbf{R}}$ and $\mathcal{L}_{\textbf{t}}$ defined using the Frobenious norm and the Huber loss (\refsec{3d_bounding_box}).
Furthermore, we consider a loss $\mathcal{L}_{\textbf{M}}$ which is similar to $\mathcal{L}_{\textbf{R}}$ but does not perform the projection into $SO(3)$, so it's not guaranteed to produce a valid rotation matrix~\cite{Levinson20arxiv}.
Finally, we compare to the rotation parameterization of \cite{Zhou19arxiv}, \ie first a classification loss $\mathcal{L}_{\text{bin}\mathbf{R}}$ over quantized bins followed by a regression loss $\mathcal{L}_{\text{off}\textbf{R}}$ to continuous offsets.
We use mean average precision (mAP) as 3D object detection metric \cite{Qi19ICCV} with 3D IoU threshold 0.25 and 0.5, as originally proposed in \cite{Song15CVPR}.
The results are shown in \reftab{ablation}. The best option is to directly predict the rotation matrix $\mathbf{R}$ using SVD and optimize it together with the translation $\mathbf{t}$ using our $\mathcal{L}_{\mathbf{Rt}}$.
%
\begin{table}[h!]
\centering
\small
\begin{tabular}{lcc}
\toprule
9-DoF Bounding Box \hspace{10px}3D mAP: & @ 0.5 & @ 0.25 \\
\midrule
$\mathcal{L}_{\text{bin}\mathbf{R}} + \mathcal{L}_{\text{off}\mathbf{R}} + \mathcal{L}_{\mathbf{t}}$ (as in \cite{Zhou19arxiv}) & 43.3 & 75.0\\ 
$\mathcal{L}_{\mathbf{M}} + \mathcal{L}_{\mathbf{t}}$ & 44.8 & 77.0\\ 
$\mathcal{L}_{\mathbf{R}} + \mathcal{L}_{\mathbf{t}}$ & 46.8 & 77.2\\ 
$\mathcal{L}_{\mathbf{Rt}}$  (\refeq{pose}, ours) & \textbf{48.6} & \textbf{77.2} \\
\bottomrule
\end{tabular}
\caption{\small\textbf{3D bounding box estimation.}
We compare different representations to estimate the rotation and translation of 3D bounding boxes.
The metric is mAP with IoU thresholds 0.5 and 0.25.}
\label{tab:ablation}
\end{table}

\paragraph{How effective is the collision loss?}
An important aspect of multiple object reconstruction is physical plausibility, \ie, reconstructed objects should not intersect.
To evaluate the effectiveness of the collision loss, we measure the mean intersecting volume (mIV) between colliding objects and the total number of collisions.
We report both metrics in \reftab{ablation_collision_loss} on the validation split of ShapeNet-triplets.
Our collision loss substantially decreases the intersecting volume and reduces the number of collisions by 60.5\%.


\begin{table}[]
\centering
\small
\begin{tabular}{lccc}
\toprule
 & mIV & \multicolumn{2}{c}{Num. Collisions}\\
\midrule
$\mathcal{L}'$ & $1168.8$ & $4116$ & \multirow{2}{*}{\ArrowDown{$-60.5 \%$}} \\
$\mathcal{L}' + \LossCollision$ (ours) & $\mathbf{794}$   & $\mathbf{1627}$ & \\ 
\bottomrule
\end{tabular}
\caption{\small\textbf{Effect of the collision loss.}
We report the mean intersection volume (mIV) over all objects and scenes, and the total number of collisions for our model with and without collision loss.}
\label{tab:ablation_collision_loss}
\end{table}

\vspace{-5px}
\paragraph{How do soft- and hard-labels affect shape estimation?}
In \refsec{shape_reconstruction}, we present two approaches to select shape exemplars from the database $\mathcal{Z}$.
The first one optimizes $\mathcal{L}'_{\textbf{z}}$ (\refeq{shape_hard}) using hard-labels, \ie one-hot encoding of target labels $z$, as done in \cite{Tatarchenko19CVPR}.
The second approach $\mathcal{L}_{\textbf{z}}$ (\refeq{shape_soft}) relies on soft-labels taking into consideration geometric similarity between objects,
therefore allowing to predict multiple plausible shapes instead of forcing the network to make a hard decision on one particular shape.
Using the evaluation methodology from \cite{Popov20ECCV}, we evaluate shape reconstruction as intersection-over-union (IoU) on a $128^3$ voxel grid (\reftab{soft_hard_labels}). We report both mean IoU over all classes and class-agnostic global IoU.
Our shape-selection mechanism using soft-labels significantly improves shape prediction by $+4.2$ mIoU over the hard-labels baseline \cite{Tatarchenko19CVPR}.

\begin{table}[]
\centering
\small
\begin{tabular}{lcc}
\toprule
Shape Estimation \hspace{15px} Abs. 3D IoU: & mean & global\\
\midrule
$\mathcal{L}'_{\mathbf{z}}$ (\refeq{shape_hard}) Hard-Labels (as in \cite{Tatarchenko19CVPR})	& $32.2$ & $40.3$\\	
$\mathcal{L}_{\mathbf{z}}$ (\refeq{shape_soft}) Soft-Labels (ours) 	& $\mathbf{36.4}$	& $\mathbf{44.7}$	 \\
\bottomrule
\end{tabular}
\caption{\small\textbf{Soft vs. hard labels.} Shape reconstruction quality in terms of intersection-over-union (IoU) on a $128^3$ voxel grid.}
\label{tab:soft_hard_labels}
\end{table}

\paragraph{Comparison to CoReNet on their datasets and Pix3D}
First, we compare our reconstructions to CoReNet~\cite{Popov20ECCV} on their ShapeNet-pairs and ShapeNet-triplets datasets.
Given an image, \cite{Popov20ECCV} predicts a dense $128^3$ voxel grid.
Each voxel is either empty or assigned to an object-class, trained with the focal loss ($m_8$)~\circlenum{1} or the IoU loss ($m_9$) \circlenum{2}, see \reftab{corenet_triplet}.
Our method reaches a higher relative 3D IoU (59.5 \vs 43.9) but does not quite match CoReNet's absolute 3D IoU (36.4 \vs 43.9).
The relative score takes the maximum possible score into account, \ie
as our model is supervised with clustered shapes (from the shape database $\mathcal{Z}$) it can only be as good as this supervision.
The oracle \circlenum{5} indicates this best possible score for our model, using the ground truth 9-DoF bounding box and the ground truth shapes from $\mathcal{Z}$ used to supervise our model.
We also perform Procrustes alignment\circlenum{4} to the ground truth to abstract from 9-DoF estimation errors  (48\% \vs 36\%).

Next, we analyze the generalization capabilities of both models under varying number of objects and class-type combinations (\reftab{corenet_num_objects}).
We train on ShapeNet-pairs and evaluate on ShapeNet-triplets, and vice-versa.
Our model generalizes well when trained on triplets and evaluated on pairs (36.41 \vs 36.21).
Both CoReNet and ours experience performance drops when trained on pairs and evaluated on triplets, but we lose less than CoReNet (-10\% \vs -22\%). 

Finally, we compare to CoReNet quantitatively on Pix3D in the same setting as \cite{Popov20ECCV}.
We report mIoU over all 9 classes and splits $S_1,S_2$ as defined by \cite{Gkioxari2019MeshR}.
On $S_1$, we obtain $34.1\%$\,(\vs$33.3\%$).
On $S_2$, $26.3\%$\,(\vs$23.6\%$).
Thus, our approach improves over CoReNet on real images.


\begin{table}
\centering
\small
\begin{tabular}{l|cc|cl}
\toprule
Method   & Train & Test &  \multicolumn{2}{c}{3D mIoU}  \\
\midrule
CoReNet \cite{Popov20ECCV} & triplets & pairs & $-$ &\\
CoReNet \cite{Popov20ECCV} & triplets & triplets & $43.9$ &\multirow{2}{*}{\ArrowDown{$-22.3 \%$}}\hspace{-20px} \\
CoReNet \cite{Popov20ECCV} & pairs & triplets & $34.1$ & \\
\midrule
Points2Objects \footnotesize{(Ours)}\hspace{-2px} & triplets & pairs & $36.2$ \\  
Points2Objects \footnotesize{(Ours)}\hspace{-2px} & triplets & triplets & $36.4$ & \multirow{2}{*}{\ArrowDown{$\mathbf{- 10.1\%}$}}\hspace{-20px}\\
Points2Objects \footnotesize{(Ours)}\hspace{-2px} & pairs & triplets & $32.7$ & \\

\bottomrule
\end{tabular}
\caption{Generalization to varying object types and cardinality.}
\label{tab:corenet_num_objects}
\end{table}


\section{Conclusion}
We have presented an end-to-end trainable model for realistic and joint 3D multi object reconstruction from a single input RGB image.
Specifically, we extend the CenterNet paradigm to coherently predict multiple 3D objects.
Objects are first detected as points, then reconstructed by jointly estimating 9-DoF object bounding boxes and 3D shape exemplars from a given shape database.
Our model is agnostic to shape representations and flexible towards changing them in the shape database.
We further aim towards realistic and physically plausible reconstructed scenes.
To that end, the model encourages collision-free reconstructions and uses CAD models as shape representations to guarantee valid and realistic object shapes.

{ \paragraph{Acknowledgments:} We thank Sergi~Caelles, Stefan~Popov and Kevis-Kokitsi Maninis for helpful discussions, 
Jonas Schult and Theodora Kontogianni for feedback on the paper and contributions to the supplementary.
Bastian Leibe's research is supported by ERC Consolidator Grant DeeViSe (ERC-2017-CoG-773161).}

\appendix
\section{Supplementary Material}

\newcommand\fordir[2]{
\begin{figure*}
\center
\includegraphics[width=0.3\textwidth]{figures/real_images/input/IMG_#1.jpg}
\includegraphics[width=0.3\textwidth, trim=250 0 250 0, clip]{figures/real_images/corenet/#1}
\scalebox{-1}[1]{\includegraphics[width=0.3\textwidth]{figures/real_images/ours/IMG_#1.png}}
\vspace{-40px}
\begin{tabular}{ccc}
    Input Image & CoReNet & Ours\\
    \hspace{0.28\textwidth} &
    \hspace{0.28\textwidth} &
    \hspace{0.28\textwidth} \\
\end{tabular}
\end{figure*}
}

In this supplementary material, we provide additional details on the inference phase of our model, as well as runtime information and we also show additional qualitative results of our approach and of CoReNet applied to real-world images recorded using a consumer mobile device.
Finally, we discuss differences and limitations of both models.

\section{Details on inference phase.}
During training, the model heatmaps $\hat{Y}$ are solely supervised at ground truth pixel locations $\mathbf{p}$, all other pixels are ignored.
The predicted heatmap values $\hat{Y}_{\mathbf{\hat{p}}_i}$ can directly be interpreted as confidence scores $s_i$.
The details of the training process are described in Sec.\,3.1 of the paper and are analogous to \cite{Zhou19arxiv}.
At test time, each pixel in the predicted heatmaps $\hat{Y} \in [0, 1]^{\frac{W}{R} \times \frac{W}{R} \times C}$ corresponds to a potential object prediction.
In our experiments, the heatmaps have a resolution of $64 \times 64$.
However, we only consider detections with a confidence score $s_i$ above a fixed threshold $\tau$, \ie, $s_i > \tau$. For the experiments on the synthetic CoReNet~\cite{Popov20ECCV} datasets, we set $\tau = 10^{-2}$ and for the inference on the real-world images (discussed below), we set $\tau = 10^{-1}$ to compensate for the larger uncertainty originating from the generalization gap due to evaluating on real data while training on synthetic data.

\section{Inference speed.}
One full pass of the presented method from the input image of size $256 \times 256$\,px$^2$ to the multi-object 3D reconstruction takes about $29$\,ms. This corresponds to \textbf{34 FPS} such that our method is real-time capable running an a P100 GPU.
Unlike \cite{Popov20ECCV}, our method does not require additional post-processing steps (\eg, marching cubes) to extract the final object surface. Instead, we place the detected shape exemplar from the shape database into the scene and align it to the predicted 9-DoF bounding box.

\newpage
\section{Additional qualitative results on real scenes.}
We show additional qualitative result on real pictures taken with a mobile phone.
The samples in this real dataset show comparable object classes and imitate the general placement of objects of the synthetic dataset which is used for training.
Besides cropping and resizing the input images to $256 \times 256$ pixels, no additional pre-processing steps are performed.
As in the main paper, our model is trained on the synthetic ShapeNet-triplet dataset presented in \cite{Popov20ECCV} and evaluated on the real images. We also show results on the CoReNet $m_9$ model, trained an the same synthetic dataset and evaluated on the newly recorded real images.
Qualitative results on both models are shown below.

\section{Discussion and comparison to CoReNet.}
In general, the method presented in this work generalizes notably better to real images than CoReNet, in terms of qualitative reconstructions. This is mostly due to the fact that we represent objects as points and formulate reconstruction as a shape selection problem. In particular, this means that our reconstructions are always valid shapes without holes and detached components.
Our strong data augmentation is likely to contribute as well.
Further, this method performs reconstruction on an object level, such that different instances of the same semantic class are represented individually along with a 9-DoF bounding box each.
In CoReNet, individual objects of the same semantic class share a dense voxel-grid.
Separating different instances could be achieved via connected components but might fail in case of wrongly disconnected parts.
On a technical side, CoReNet predicts a dense voxel grid aligned to the image.
This leads to scalability issues with increasing scene size and it cannot reconstruct objects that are truncated in the input image.
Finally, CoReNet reconstructs objects in the camera coordinate system while our method predicts objects in the world-coordinate system. Our approach has the advantage that objects can directly be rendered on a ground plane. However, both methods perform the reconstruction only up to scale due to missing camera extrinsics at test.
\fordir{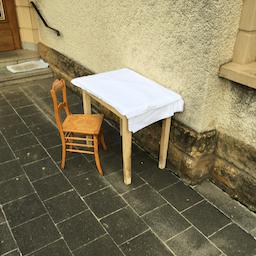}{''}
\fordir{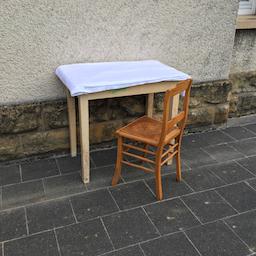}{''}
\fordir{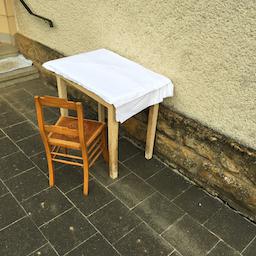}{''}
\fordir{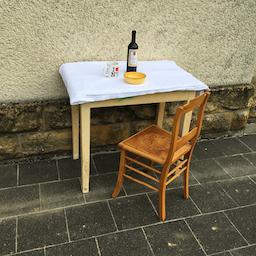}{''}
\fordir{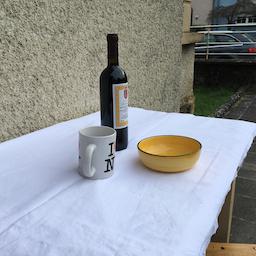}{''}
\fordir{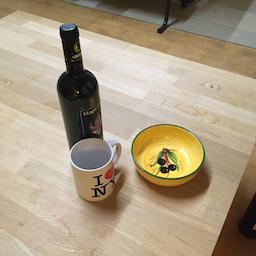}{''}
\fordir{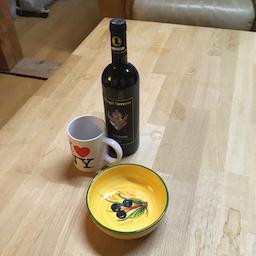}{''}
\fordir{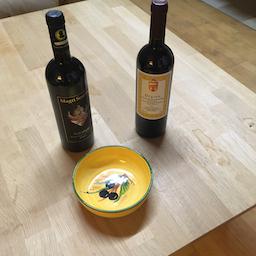}{''}
\fordir{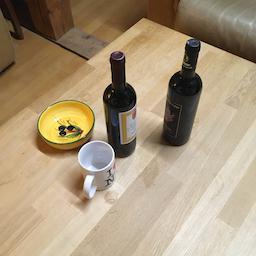}{''}
\fordir{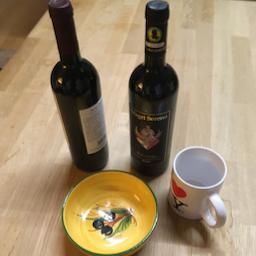}{''}
\fordir{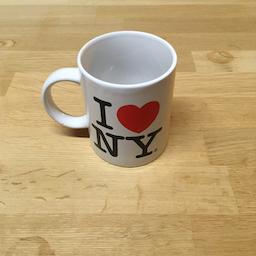}{''}
\fordir{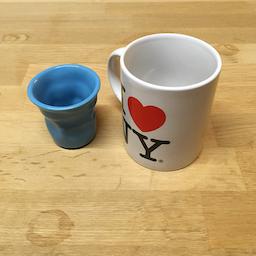}{''}

\clearpage
\clearpage
{\small
\bibliographystyle{ieee_fullname}
\bibliography{abbreviations,egbib}
}

\end{document}